\documentclass[10pt,twocolumn,letterpaper]{article}

\usepackage{icb}
\usepackage{times}
\usepackage{epsfig}
\usepackage{graphicx}
\usepackage[caption=false,font=footnotesize]{subfig}
\usepackage{hyperref}
\usepackage{amsmath}
\usepackage{amssymb}
\usepackage{float}
\usepackage{comment}
\usepackage{booktabs}
\usepackage{array}
\usepackage{multicol}

\hypersetup{
  colorlinks, linkcolor=red
}

\makeatletter
\g@addto@macro{\UrlBreaks}{\UrlOrds}
\makeatother

\newcolumntype{M}[1]{>{\centering\arraybackslash}m{#1}}

\icbfinalcopy 

\ificbfinal\fi
\begin{document}

\title{Matching Fingerphotos to Slap Fingerprint Images\thanks{This research is supported by caribou Digital who are funded by the Bill and Melinda Gates Foundation.}}

\author{Debayan Deb$^1$, Tarang Chugh$^1$, Joshua Engelsma$^1$, Kai Cao$^1$,\\Neeta Nain$^2$, Jake Kendall$^3$, and Anil K. Jain$^1$\\
$^1$Department of Computer Science and Engineering, Michigan State University, East Lansing, MI 48824\\
$^2$ Malaviya National Institute of Technology, Jaipur, India\\
$^3$ Digital Financial Services Lab, Caribou Digital, UK\\
{\small E-mail: $^1$\tt{\{debdebay,chughtar,engelsm7,kaicao,jain\}@cse.msu.edu},}\\
{\small$^2$\tt{nnain.cse@mnit.ac.in}, $^3$\tt{jake@cariboudigital.net}}
}

\maketitle
\thispagestyle{empty}


\begin{abstract}
We address the problem of comparing fingerphotos, fingerprint images from a commodity smartphone camera, with the corresponding legacy slap contact-based fingerprint images. Development of robust versions of these technologies would enable the use of the billions of standard Android phones as biometric readers through a simple software download, dramatically lowering the cost and complexity of deployment relative to using a separate fingerprint reader. Two fingerphoto apps running on Android phones and an optical slap reader were utilized for fingerprint collection of $309$ subjects who primarily work as construction workers, farmers, and domestic helpers. Experimental results show that a True Accept Rate (TAR) of $95.79$ at a False Accept Rate (FAR) of $0.1\%$ can be achieved in matching fingerphotos to slaps (two thumbs and two index fingers) using a COTS fingerprint matcher. By comparison, a baseline TAR of $98.55\%$ at $0.1\%$ FAR is achieved when matching fingerprint images from two different contact-based optical readers. We also report the usability of the two smartphone apps, in terms of failure to acquire rate and fingerprint acquisition time. Our results show that fingerphotos are promising to authenticate individuals (against a national ID database) for banking, welfare distribution, and healthcare applications in developing countries.
\end{abstract}

\section{Introduction}

\begin{figure}[t!]
    \centering
    \includegraphics[clip, trim=0cm 0.25cm 0cm 0cm, width=0.88\linewidth]{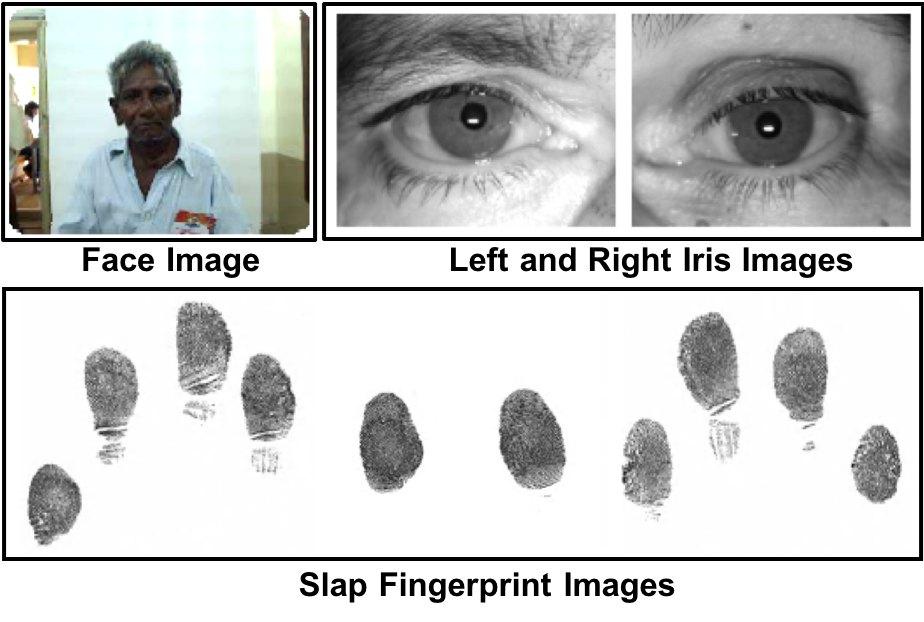}
    \caption{India's national ID program, Aadhaar, captures a face image, left and right iris images, and slap (4-4-2) fingerprint images, to enroll its residents and assign them a 12-digit unique identifier~\cite{uidai}.}
    \label{fig:aadhaar1}
\end{figure}

\begin{figure}[t!]
    \centering
    \subfloat[]{\includegraphics[width=0.45\linewidth]{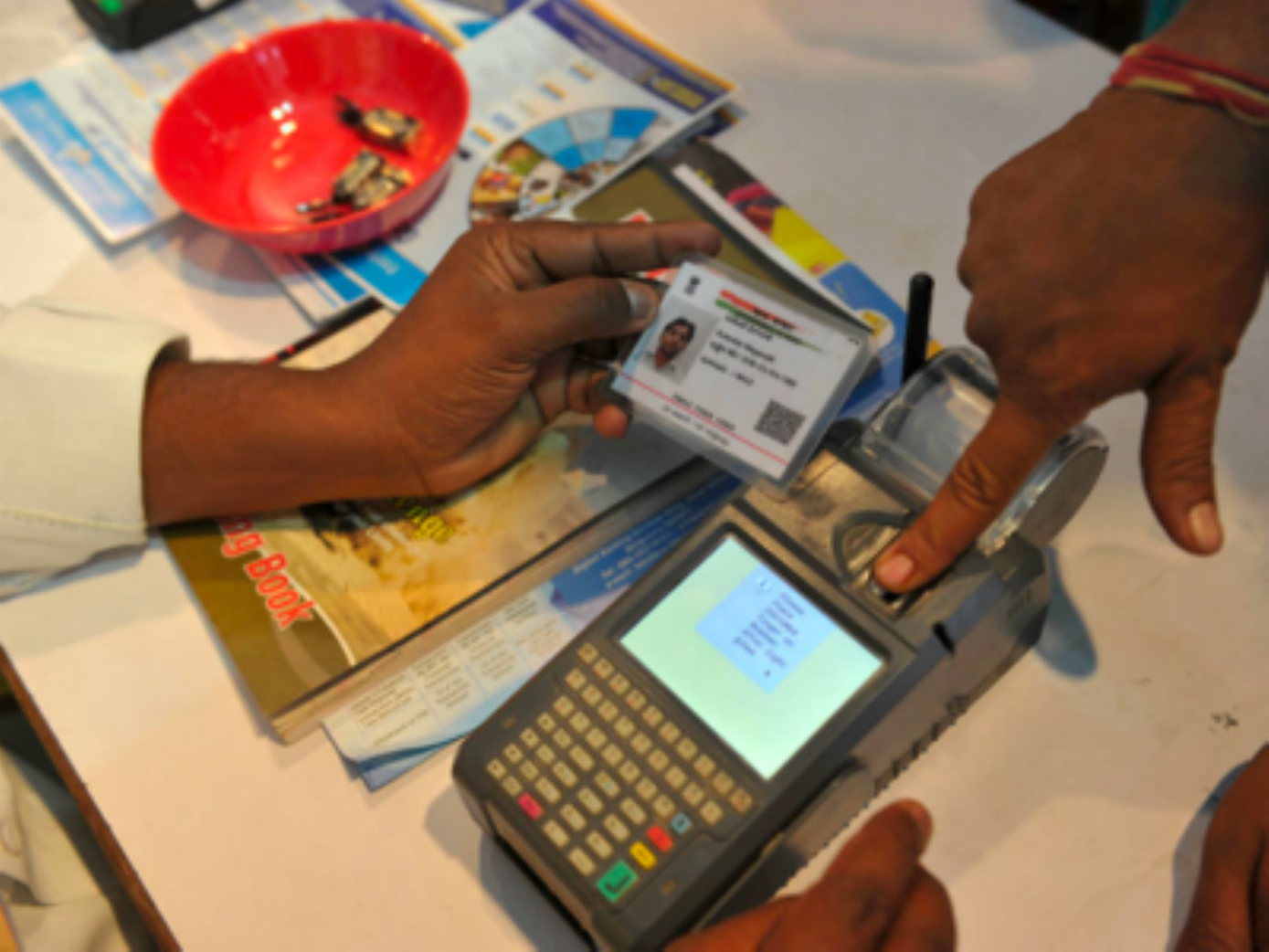}}\hfil
    \subfloat[]{\includegraphics[width=0.49\linewidth]{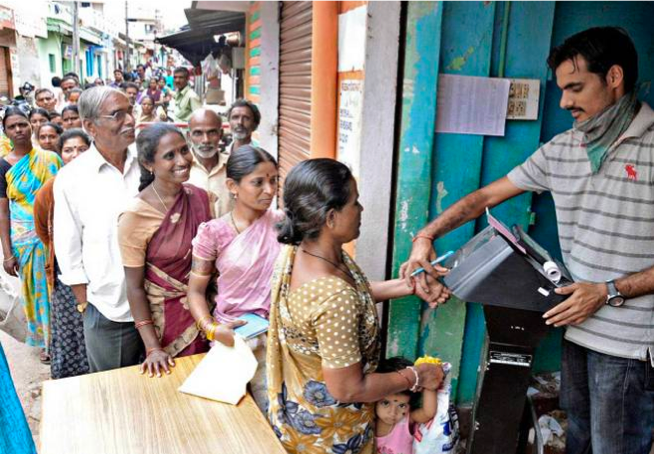}}
    \vspace{1mm}
    \caption{Contact-based optical fingerprint readers used at benefit distribution centers for user authentication~\cite{uidai}.}
    \label{fig:aadhaar2}
\end{figure}

\begin{table*}[!t]
    \centering
    \caption{Related work on smartphone camera based fingerprint authentication.}
    \resizebox{\linewidth}{!}{
    \begin{tabular}{M{2.2cm}M{3.1cm}M{6.2cm}M{2.4cm}M{6.2cm}}
    \toprule
        \textbf{Study} & \textbf{Fingerprint Capture} & \textbf{Objective} & \textbf{Database} & \textbf{Comments} \\ 
        \midrule
         Derawi et al., 2011~\cite{derawi2011fingerprint} & Smartphone Camera, \newline 2 different models &  Fingerphoto-to-fingerphoto matching; No preprocessing & 22 subjects; 1,320 images & Neurotechnology VeriFinger 6.0 SDK~\cite{veri} used with RGB images; EER = 4.5\% \\ \midrule
         
         Stein et al., 2012~\cite{stein2012fingerphoto} & Smartphone Camera, \newline 2 different models & Automatic fingerprint segmentation and on-device matching & 41 subjects; \newline 656 images & Controlled capture set-up with smartphone held on a stand; \newline EER = 19.1\% \\ \midrule
         
         Stein et al., 2013~\cite{stein2013video} & Smartphone Camera \newline Video Sequences, \newline 2 different models & Finger video sequences for fingerphoto-to-fingerphoto matching; anti-spoofing techniques &  37 subjects; \newline 66 finger videos; 2,100 images & Constrained background and illumination; EER = 3.0\% \\ \midrule
         
          Li et al., 2013~\cite{li2013quality} & Smartphone Camera, \newline 3 different models & Fingerphoto quality estimation; quality assessment using matching performance  & 25 subjects; 2,100 images  & Manual segmentation of finger ROI; EER ranges between 2.7\% and 35.3\% based on estimated quality levels \\  \midrule
          
         Sankaran et al., 2015~\cite{sankaran2015smartphone} & Smartphone Camera; Optical Reader & Scattering Network based feature representation & 64 subjects; 5,120 images & Limited subject diversity; EER = 3.65\% \\ \midrule
         
         \textbf{This study} & Smartphone Camera, \newline 2 different apps; \newline Optical Readers, \newline 2 different models & \textit{In-situ} evaluation of fingerphoto-to-slap-fingerprint matching; database of 309 subjects with different occupations in India & 309 subjects; 7,976 images & Innovatrics IDKit SDK~\cite{inno}; \newline TAR = 98.01\% $@$ FAR =1.0\%; baseline performance in matching fingerprint images from two different contact-based optical readers of TAR = 98.55\% $@$ FAR = 1.0\%\\ \bottomrule
    \end{tabular}}
    \label{tab:relatedwork}
\end{table*}

A large proportion of individuals, especially economically disadvantaged, in developing countries around the world often lack any type of identification documents making it difficult for them to access government benefits, healthcare, and financial services. To address this critical need, efforts are being made to build large-scale national biometric databases to efficiently and effectively authenticate individuals at the point of service. The world's largest biometric-based national ID program is the India's Aadhaar~\cite{uidai}. It has already enrolled face, fingerprints, and irides of over 1.2 billion residents\footnote{\url{https://uidai.gov.in/aadhaar_dashboard/india.php} Accessed: Apr 2, 2018}. See Fig.~\ref{fig:aadhaar1}. Given the success of national ID programs, such as India's Aadhaar, Pakistan's NADRA\footnote{\url{https://www.nadra.gov.pk/}}, and other programs, new opportunities to leverage fingerprint authentication for day-to-day transactions are rapidly becoming commonplace\footnote{\url{https://aadharpaymentapp.org/}}.

In a National ID system, such as Aadhaar, a user provides their unique 12-digit Aadhaar number along with their fingerprint (Fig.~\ref{fig:aadhaar2} (a)). Next, an encrypted template of the fingerprint image is relayed to the Aadhaar server for authentication. If authenticated, the user is eligible to receive benefits or services (Fig.~\ref{fig:aadhaar2} (b)).

Given the scale at which National ID programs function, fingerprint recognition systems need to deliver higher authentication accuracy, usability, and low-cost authentication solutions. This requires: low-cost biometric readers, fast fingerprint acquisition, integration of widely available commodity smartphones in the sensing and authentication process, and ability to process noisy fingerprints for persons engaged in manual work~\cite{maltoni2009handbook}.

\begin{figure}[t!]
    \centering
    \includegraphics[clip, trim=0cm 0.4cm 0cm 0.2cm, width=\linewidth]{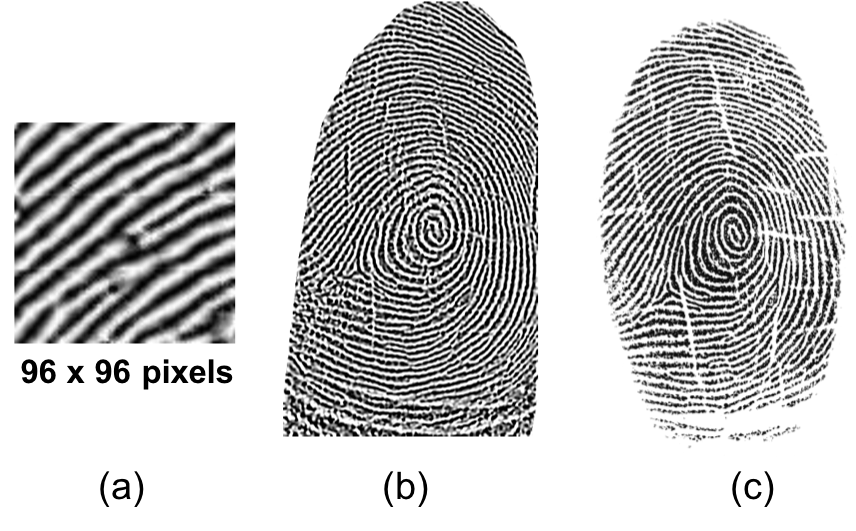}
    \vspace{-2mm}
    \caption{Fingerprint images (500 ppi) of the same finger captured using a (a) capacitive sensor embedded in a smartphone*~($96 \times 96$ pixels), (b) smartphone camera (fingerphoto captured by App2) (380 $\times$ 540 pixels), and (c) optical slap fingerprint reader (CrossMatch) (330 $\times$ 512 pixels).}
    \label{fig:capacitive}
    \raggedright{\scriptsize{*Courtesy: Shenzhen Goodix \url{http://www.goodix.com/}}}
\end{figure}


\textit{TouchID} by Apple in 2013~\cite{touchid} dramatically changed the way how we unlock our phones and use smartphones for mobile payment using our fingerprints. Since then mobile phones have also been introduced with iris and face recognition capabilities\footnote{\url{https://spectrum.ieee.org/tech-talk/consumer-electronics/gadgets/new-samsung-galaxy-s8-unlocks-with-facial-recognition-iris-scanning}}. Still, for smartphone unlock and mobile payments, fingerprints appear to be by far the most popular. However, in the case of fingerprints, smartphones must be fitted with embedded capacitive fingerprint sensors which implies only new phones can be used in this way and restricts usage of the billions of existing smartphones. Additionally, embedded capacitive sensors typically feature a small sensing area ($\sim90 \times 90$ pixels) that can capture only partial fingerprints which are not appropriate for matching with legacy slap fingerprints in the national ID databases. In addition, fingerprints acquired by the smartphones are proprietary and inaccessible even to the smartphone user. Hence, they cannot be used to authenticate the user against the national ID databases in an inter-operable fashion. For this reason, fingerphoto captured by smartphone camera is a more effective solution for user authentication. See Fig.~\ref{fig:capacitive} for a comparison of fingerprint image captured by a (a) capacitive sensor embedded in a smartphone, (b) smartphone camera, and (c) slap fingerprint reader. If a fingerphoto can be successfully compared against legacy slap fingerprints in a national ID database for authentication, it will obviate the need for a separate fingerprint reader at benefit distribution centers or Point of Sale (PoS) as shown in Fig.~\ref{fig:aadhaar2} reducing dramatically the cost and complexity of deploying and maintaining such systems. Another advantage of fingerphoto is that it is touch-less acquisition, hence, no residual fingerprint impression is left behind as in the case of touch-based sensors.

Fingerphoto based authentication can be broadly classified into two categories: (i) fingerphoto-to-fingerphoto matching, and (ii) fingerphoto-to-slap-fingerprint matching. In the first category, a \textit{fingerphoto} is matched against a previously enrolled fingerphoto from the same smartphone. This provides an alternate solution to capacitive sensors for fingerprint acquisition in smartphones. However, this application is primarily meant for smartphone unlock and possibly mobile payment re-authentication (post-enrollment) to a bank or mobile operator system. But it lacks inter-operability with legacy slap images which would typically populate e.g. a national ID database, and which might be used for initial on-boarding and bank Know-Your-Customer (KYC) compliance. Hence, fingerphoto-to-legacy slap fingerprint matching addresses a broader set of authentication problems.

\begin{figure}[t!]
  \centering
  \includegraphics[width=0.8\linewidth]{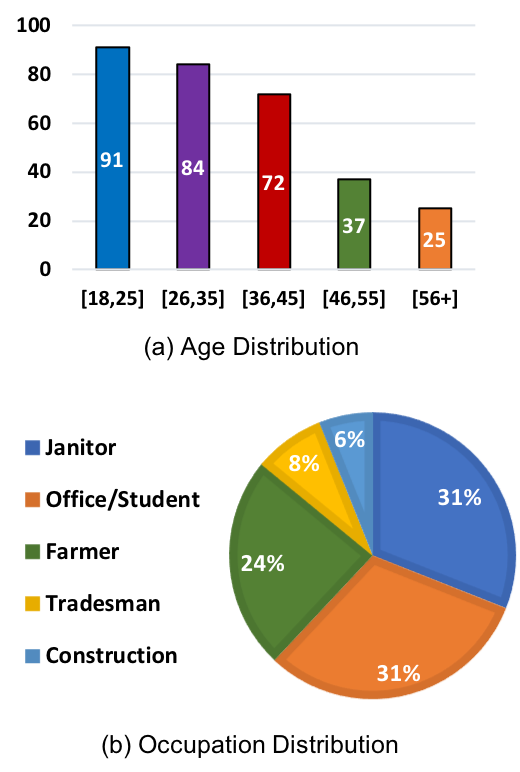}
   \caption{Demographics of 309 subjects who provided their fingerprints in our study. (a) Age (in yrs.), and (b) occupation. 57\% of the subjects were males and the remaining 43\% were females.}
  \label{fig:demographics}
\end{figure}

Our objective is to determine whether fingerphotos of sufficiently high resolution and fidelity\footnote{Jan Krissler, a German hacker known as StarBug, used high resolution images captured by a digital single-lens reflex (DSLR) camera, from a distance of 3 meters to recreate the fingerprint ridge structure of German Defense Minister, Ursula von der Leyen. \url{https://www.theguardian.com/technology/2014/dec/30/hacker-fakes-german-ministers-fingerprints-using-photos-of-her-hands}} can be used to authenticate individuals who were enrolled using optical slap fingerprint readers.

In literature, both fingerphoto-to-fingerphoto and fingerphoto-to-fingerprint matching protocols have been suggested. Derawi~\textit{et al.}~\cite{derawi2011fingerprint} utilized two smartphones, Nokia N95 and HTC Desire, to capture 1,320 fingerphotos of 22 subjects. A commercial fingerprint feature extractor and matcher was employed for user authentication. Some studies have proposed pre-processing algorithms to enhance the performance of fingerphoto-to-fingerphoto matching~\cite{stein2012fingerphoto},~\cite{stein2013video},~\cite{sankaran2015smartphone}. Stein~\textit{et al.} proposed fingerphoto matching using video sequences, instead of static RGB fingerprint images, with uniform background and illumination~\cite{stein2013video}. Sankaran~\textit{et al.}~\cite{sankaran2015smartphone} and Gupta~\textit{et al.}~\cite{gupta2017fingerprint} utilized Scattering Network and Gabor Filters, respectively, for feature representations in the fingerphoto-to-fingerprint matching scenario. Stein~\textit{et al.}~\cite{stein2013video} and Taneja~\textit{et al.}~\cite{taneja2016fingerphoto} have explored the fingerprint anti-spoofing techniques for smartphone based authentication. However, these prior studies on fingerphoto based authentication are limited in scope due to (i) data collection in constrained settings, (ii) lack of representative subject demographics, and (iii) small number of subjects. Table~\ref{tab:relatedwork} summarizes and compares these studies with our study. There also has been some prior work on smartphone camera based palmprint recognition~\cite{aoyama2013contactless, moco2014smartphone, methani2010camera}.

\begin{figure}[t!]
  \centering
  \includegraphics[width=\linewidth]{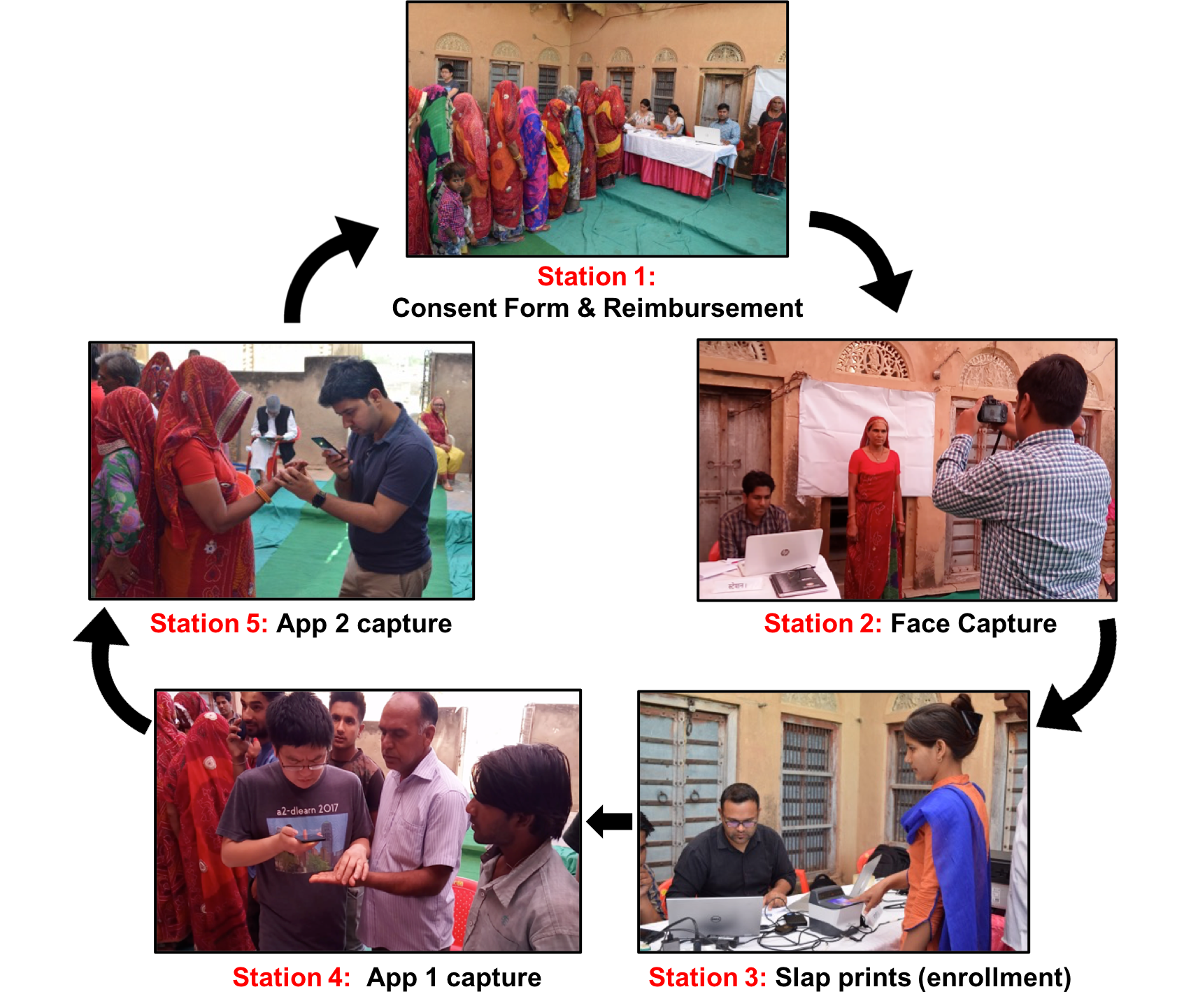}
  \caption{Process flow for the data collection process.}
  \label{fig:data_collection0}
\end{figure}

\begin{figure}[t!]
  \centering
  \includegraphics[width=0.71\linewidth]{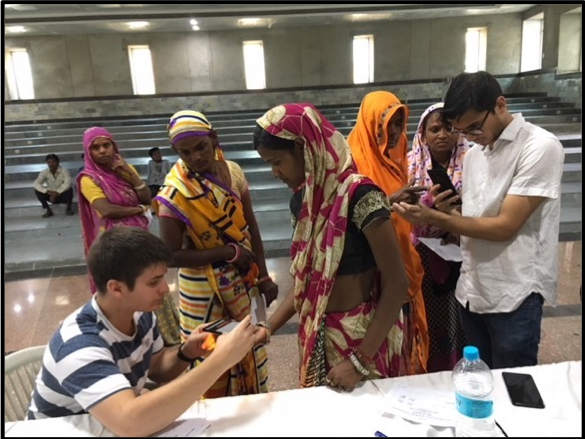}
  \caption{Data collection at MNIT, Jaipur, India, inside a lecture theatre with typical indoor illumination conditions.}
  \label{fig:data_collection1}
\end{figure}

\begin{figure}[t!]
  \centering
  \includegraphics[width=0.76\linewidth]{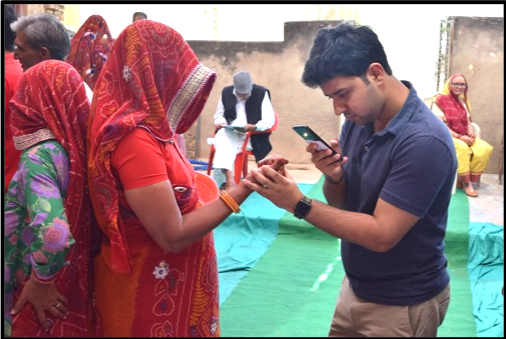}
  \caption{Outdoors data collection environment in a courtyard covered with a canopy in the village of Jhunjhunu, Rajasthan, India.}
  \label{fig:data_collection2}
\end{figure}

User authentication in large national ID programs, such as India's Aadhaar and Pakistan's NADRA, requires (i) high accuracy (e.g., FAR = 0.1\% @ FRR = 2.0\%), (ii) high usability, (iii) high throughput, (iv) low cost, and (v) low failure to acquire rate. Amid the growing concerns on user privacy, one of the major issues facing these national ID programs is the exclusion of people from receiving their benefits for individuals with poor quality fingerprints. According to NY Times, one recent study\footnote{\url{https://www.nytimes.com/2018/04/07/technology/india-id-aadhaar.html}} found that ``20 percent of the households in Jharkand state had failed to get their food rations under Aadhaar-based (fingerprint) verification". Another study found that 50,151 out of 85,589 surveyed welfare beneficiaries failed to access daily rations from 125 stores due to high false reject rates by fingerprints~\cite{hindustan}. This is true in many rural areas of India, where most of the population is involved in manual work and have worn-out or damaged fingers. We believe, a fingerphoto based authentication is a plausible solution to authenticate individuals with different occupations in a developing country setting. With this objective, we evaluate two fingerphoto solutions developed in the form of two Android apps\footnote{These applications were developed by two vendors funded by Caribou Digital via their grant from the Bill and Melinda Gates Foundation.} which we call as App1 and App2. Each App has been developed by a different commercial biometrics technology company.

\section{Data Collection}
A total of 309 subjects, above the age of 18 years, were enlisted for data collection in our study. Among these, 57\% of the subjects were males and the remaining 43\% were females. Fig.~\ref{fig:demographics} shows the age and occupation distributions of the subjects. The data collection was conducted in India at two different locations: (i) an indoor lecture theater of of MNIT, Jaipur, India, with typical indoor illumination (200 subjects), and (ii) an outdoor courtyard covered with a canopy in the village of Jhunjhunu, India, with natural lighting (109 subjects). In order to complete all the data collection in 5 working days, the enrollment and verification data was collected in the same session but at different stations. See Figs.~\ref{fig:data_collection0},~\ref{fig:data_collection1}~and~\ref{fig:data_collection2} for an illustration of process flow and data capture environments.

\begin{figure}[t!]
  \centering
  \subfloat[]{\includegraphics[width=0.6\linewidth]{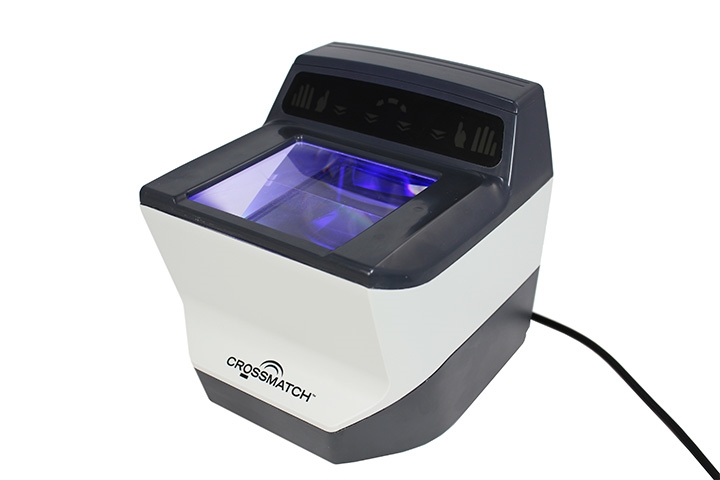}}
  \subfloat[]{\includegraphics[width=0.4\linewidth]{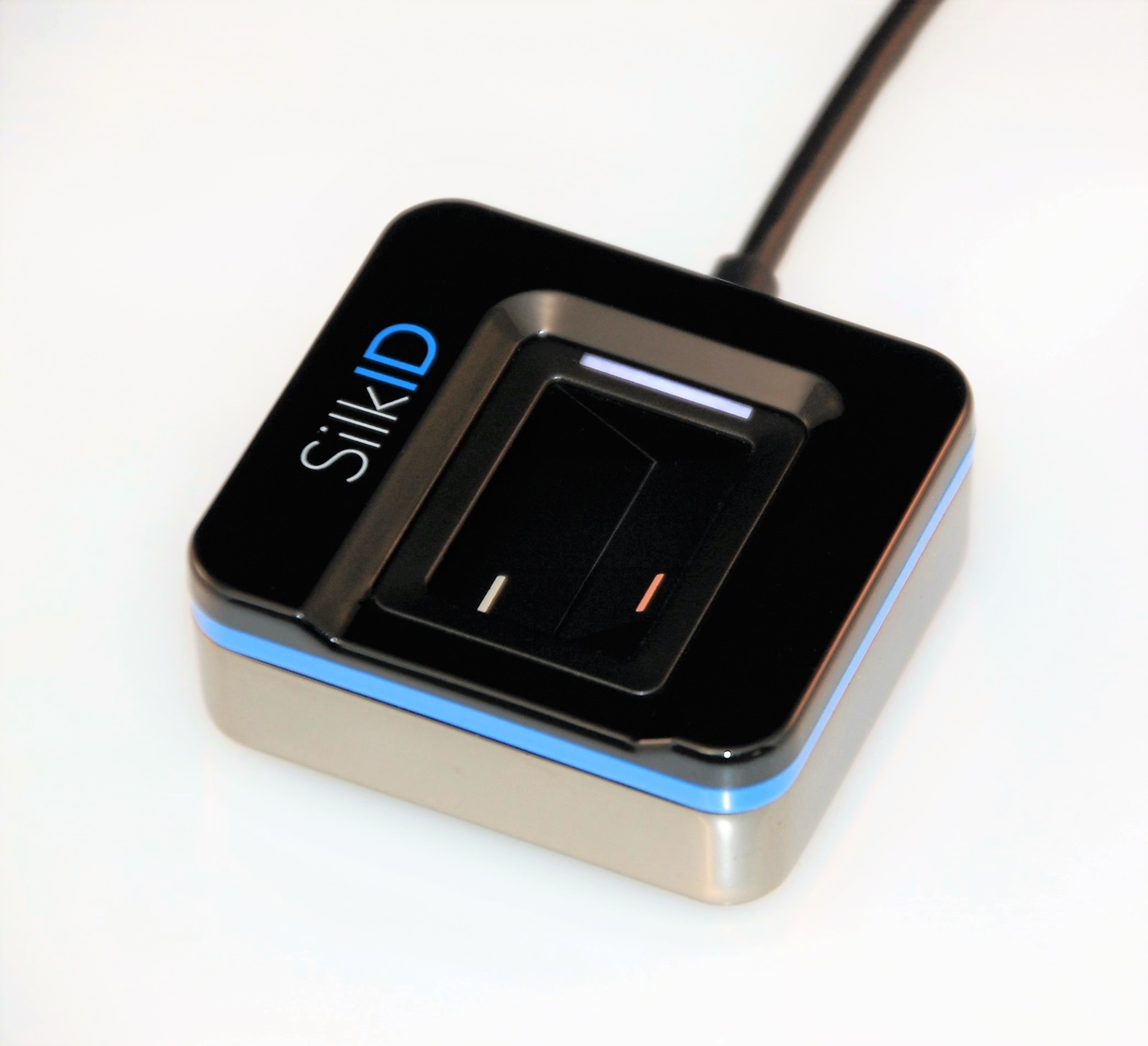}}
   \caption{Contact-based optical fingerprint readers (500ppi) for enrollment. (a)~CrossMatch Guardian 200 slap reader for enrollment~\cite{crossmatch}; (b)~SilkID (SLK20R) fingerprint reader for establishing the baseline~\cite{silkid}.}
  \label{fig:crossmatch}
\end{figure}

\begin{figure}[t!]
  \centering
  {\includegraphics[width=0.7\linewidth]{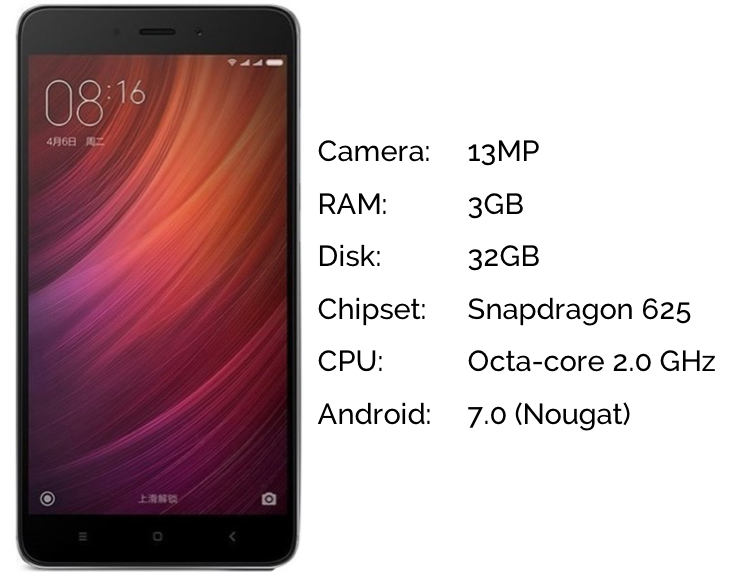}}\\
   \caption{Xiaomi Redmi Note 4 smartphone, which costs less than \$150, was utilized for fingerphoto collection.~\cite{xiaomi}.}
  \label{fig:xiaomi}
\end{figure}

\begin{figure}[t!]
  \centering
  \includegraphics[width=\linewidth]{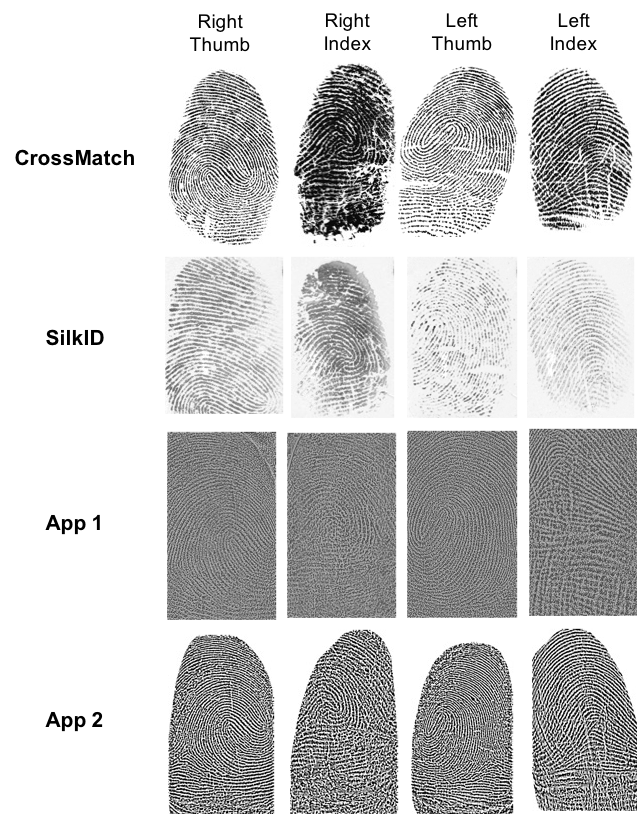}
  \caption{Comparison of fingerprint images from contact-based fingerprint readers (CrossMatch and SilkID) and the corresponding fingerphotos from App1 and App2.}
  \label{fig:collage}
\end{figure}

\begin{figure*}[t!]
  \centering
  \includegraphics[width=0.94\linewidth]{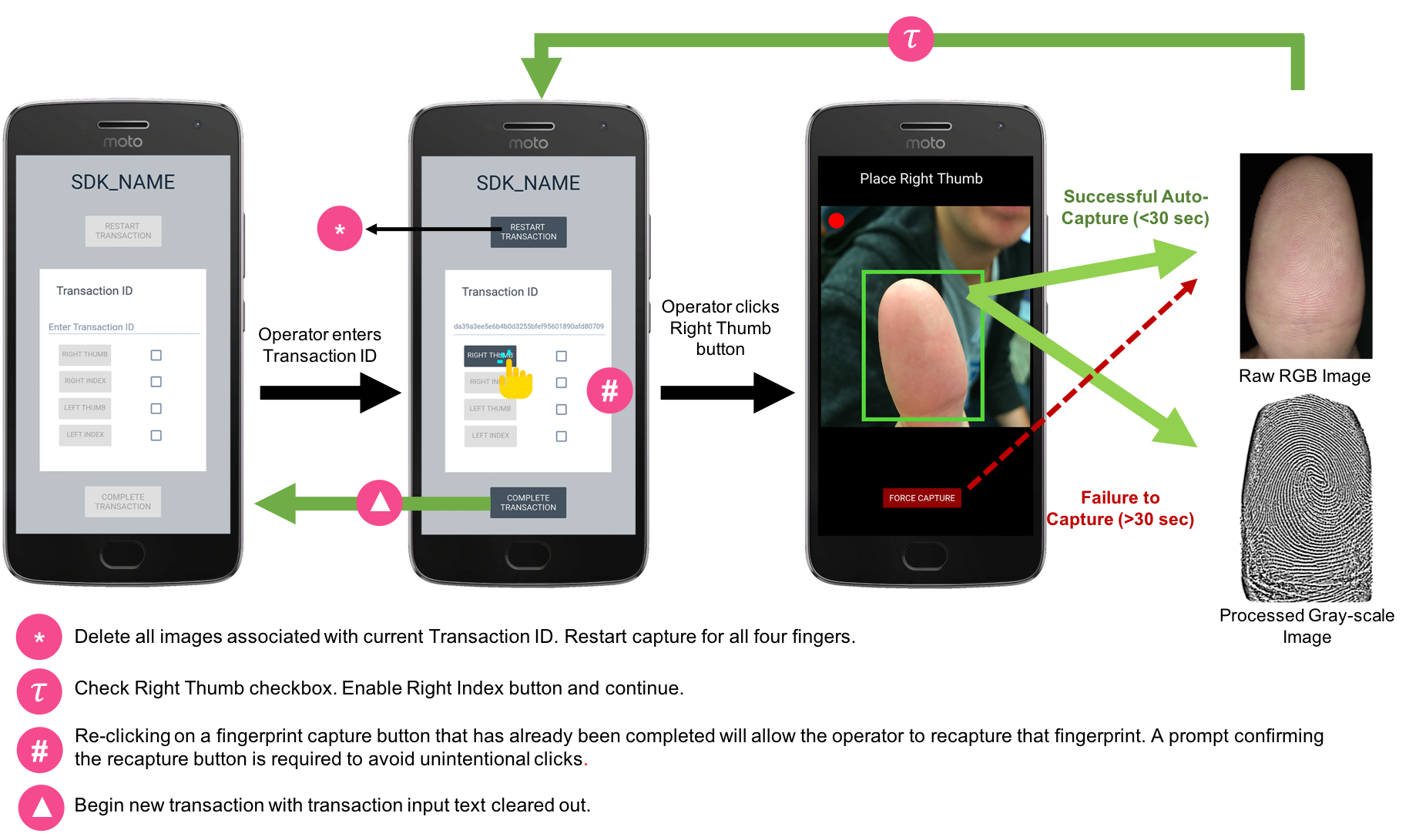}
  \caption{Schematic of the fingerprint capture process on Android apps. The two apps output both raw and processed gray-scale fingerphotos.}
  \label{fig:flow}
\end{figure*}

Fingerprint images of the two thumbs and two index fingers were first enrolled using an optical slap reader, \textit{CrossMatch Guardian 200}\footnote{CrossMatch Guardian 200 optical fingerprint reader is commonly used for slap fingerprint acquisition in National ID, Law Enforcement, and Homeland Security applications.}~\cite{crossmatch} (see Fig.~\ref{fig:crossmatch}) for a total of 2,472 enrollment images (309 subjects $\times$ 4 fingers $\times$ 2 impressions/finger). For verification, the two smartphone based Android apps were used to capture two impressions of the four enrolled fingers (two thumbs and two index fingers) from the same set of 309 subjects. Both the apps were installed on the same smartphone device, Xiaomi Redmi Note 4 (32GB model)~\cite{xiaomi} that costs less than \$150. See Fig.~\ref{fig:xiaomi}. This smartphone model was selected as it provided the best specifications at an affordable price and is one of the best-selling devices on a popular e-commerce site in India. Other Android smartphones such as Lenovo K8 Plus\footnote{\url{https://www.amazon.in/Lenovo-k8-LENOVO-Venom-Black/dp/B075HJD15N}} or Micromax Canvas Infinity\footnote{\url{https://www.amazon.in/Micromax-Canvas-Infinity-Black-18/dp/B0725RBY9X}} that cost less than \$150 can also be utilized\footnote{The two apps require a minimum camera resolution of 5 Megapixels.}. The two apps output both raw RGB fingerphoto and the corresponding processed grayscale fingerprint image. Figure~\ref{fig:collage} presents fingerprint images from contact-based fingerprint readers (CrossMatch and SilkID) and the corresponding fingerphotos from App1 and App2. 

\emph{App Design:}
A user-friendly app design that allows high throughput and prevents human error was provided to the two vendors for designing their solutions. The proposed design enabled the apps to (i) initiate a new data capture transaction that links all collected fingerphotos to unique IDs, (ii) track and streamline the data capture process, (iii) recapture any required fingerphotos, and (iv) restart an on-going transaction that safely deletes all images associated with the current transaction avoiding orphan images in storage. Fig.~\ref{fig:flow} demonstrates the flow of the fingerprint capture process on Android apps, and presents a sample pair of raw and the corresponding processed fingerphoto images captured by one of the apps. A common specification was chosen so that differences in the user interface specification would not confound the performance tests.

Given the fingerprint or fingerphoto, ISO (ISO/IEC19794-2:2011)~\cite{iso} templates are extracted from them. Comparison scores (between slap fingerprints and fingerphotos) were generated using a COTS software\footnote{Innovatrics IDKit SDK~\cite{inno}}. To obtain a baseline performance, we also acquired two impressions of the two thumbs and two index fingers for 70 subjects, for a total of 560 fingerprint images, using \textit{SilkID} (SLK20R) optical fingerprint reader~\cite{silkid}.


In addition to fingerprint matching performance, the following metrics were also logged from the two apps: (i) acquisition time per fingerphoto, and (ii) failure to acquire. These metrics are related to the usability and throughput of each app.

\section{Experimental Results}

Verification performance\footnote{National ID programs already perform de-duplication at the time of enrollment using identification (1:N) search. This study focuses on applications, such as benefit distribution and financial transactions, where a user claims an identity and is verified through their fingerphoto.} (1:1 comparison) is reported for (i) individual fingers, \textit{i.e.} each of the two thumbs and two index fingers, (ii) fusion of two fingers from the same hand, \textit{i.e.} right thumb fused with right index finger, and left thumb fused with left index finger, (iii) fusion of two fingers from different hands, \textit{i.e.} two thumbs, and two index fingers, and (iv) fusion of all four fingers, \textit{i.e.} two thumbs and two index fingers. In all of the fusion scenarios, score level fusion with simple sum rule was utilized. No score normalization was needed since the same COTS outputs all the scores in the range [0, 800]. Under verification scenario, we report True Accept Rate (TAR) @ False Accept Rate (FAR) of 0.1\%.

\begin{table}[!t]
    \centering
    \caption{TAR $@$ FAR = 0.1\% for the two fingerphoto Apps and SilkID reader for matching with enrollment fingerprint images from CrossMatch reader.}
    \vspace{1mm}
    \resizebox{0.9\linewidth}{!}{
    \begin{tabular}{lccc}
    \toprule
        \textbf{Verification experiment} & \multicolumn{3}{c}{\textbf{TAR (\%) $@$ FAR = 0.1\%}} \\ 
        \midrule
        & App1 & App2 & SilkID \\
        \toprule
        Right Thumb & 37.07 & 74.14 & 97.83 \\
        \midrule
        Left Thumb & 46.35 & 76.63 & 94.20\\
        \midrule
        Right Index & 21.17 & 67.96 & 97.46\\
        \midrule
        Left Index & 18.97 & 70.79 & 92.39\\
        \midrule
        Two Thumbs & 58.05 & 89.86 & 98.55\\
        \midrule
        Two Index fingers & 27.97 & 83.76 & 98.19\\
        \midrule
        Right Thumb and Index & 43.68 & 86.43 & 98.55\\
        \midrule
        Left Thumb and Index & 47.51 & 88.14 & 94.20\\
        \midrule
        \textbf{Fusion of all 4 fingers} & \textbf{54.98} & \textbf{95.79} & \textbf{98.55} \\
         \bottomrule
    \end{tabular}
    }
    \label{tab:results}
\end{table}

\begin{figure*}[htbp!]
    \centering
        \subfloat[App1]{\includegraphics[width=.4\linewidth]{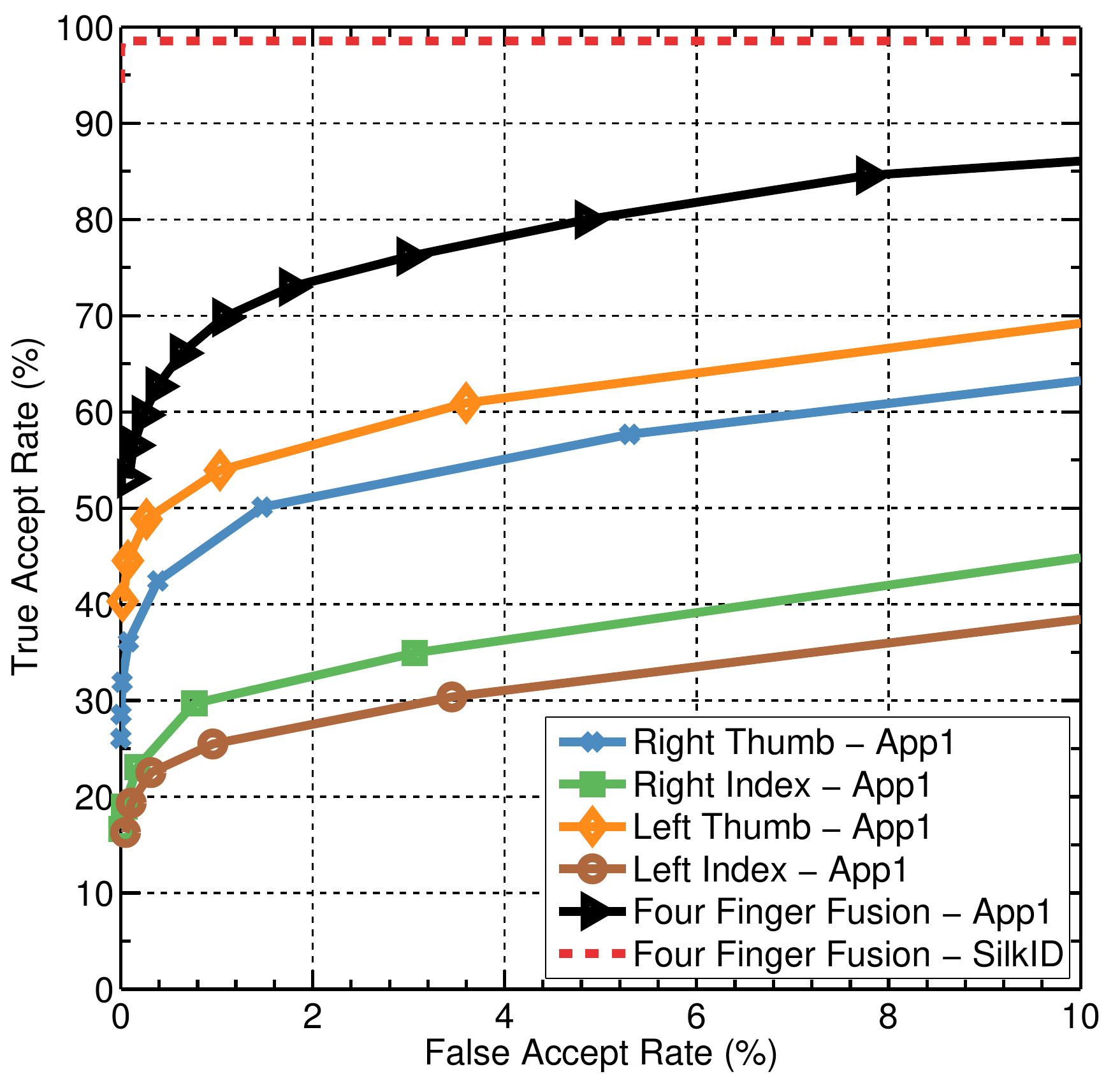}} \hfil
        \subfloat[App2]{\includegraphics[width=.4\linewidth]{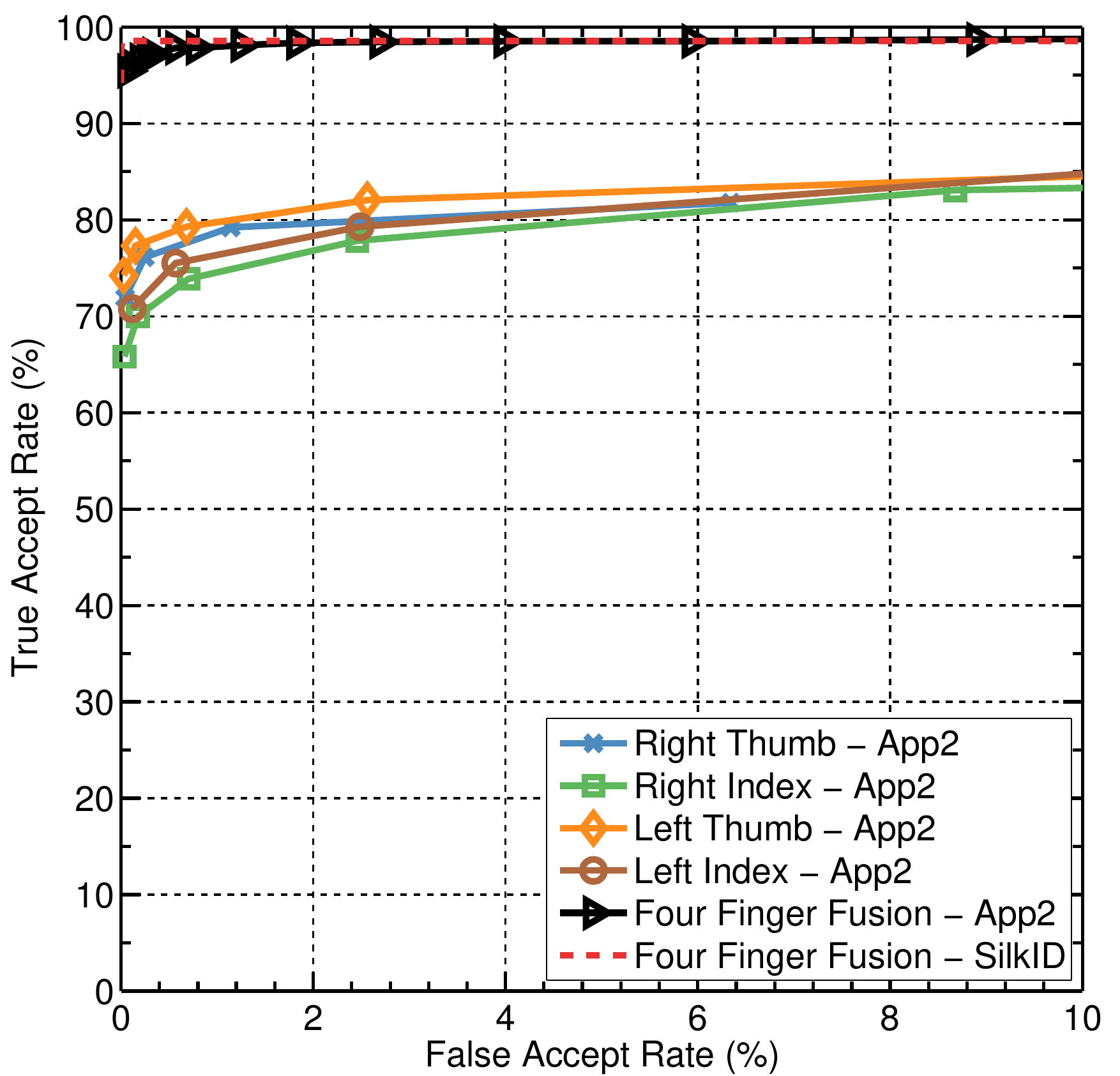}}
        \vspace{1mm}
        \caption{ROC curves for Individual Fingers for (a) App1 and (b) App2.}
        \label{fig:ind_fingers}
\end{figure*}

Fingerphotos collected by each of the two apps are compared with enrollment images from the slap scanner. In all of the matching experiments, 4,944 genuine scores (309 subjects $\times$ 4 fingers $\times$ 2 enrollment impressions $\times$ 2 verification impressions) and 95,172 impostor scores (309 fingers $\times$ 308 fingers) are computed. In order to compare the performance of the two apps, we consider the following four scenarios:

\begin{enumerate}

    \item \emph{Individual fingers}: Scores are computed separately for individual fingers (i.e. right thumb fingerphotos are only compared to right thumb fingerprints and right index fingerphotos are only compared to right index fingerprints). See Table~\ref{tab:results} and Figure~\ref{fig:ind_fingers}. 
    
    \item \emph{Fusion of two thumb (index fingers) scores}: Scores from the two thumbs (index fingers) are fused using the sum rule (i.e. right thumb score is fused with left thumb score; right index finger score is fused with left index finger score). See Table~\ref{tab:results}.
    
    
    
    \item \emph{Fusion of right hand (left hand) thumb and index finger}: Scores from the right thumb and right index (or left thumb and left index) are fused by adding both scores together. See Table~\ref{tab:results}.
    
    
    
    \item \emph{Fusion of scores from all four fingers (two thumbs and two index fingers)}: Scores from all fingers (two thumbs and two indexes) are fused by adding all four scores together (i.e. right thumb, right index, left thumb and left index scores). See Table~\ref{tab:results} and Figure~\ref{fig:ind_fingers}.
    
\end{enumerate}

\begin{figure}[t!]
    \centering
        \includegraphics[width=\linewidth]{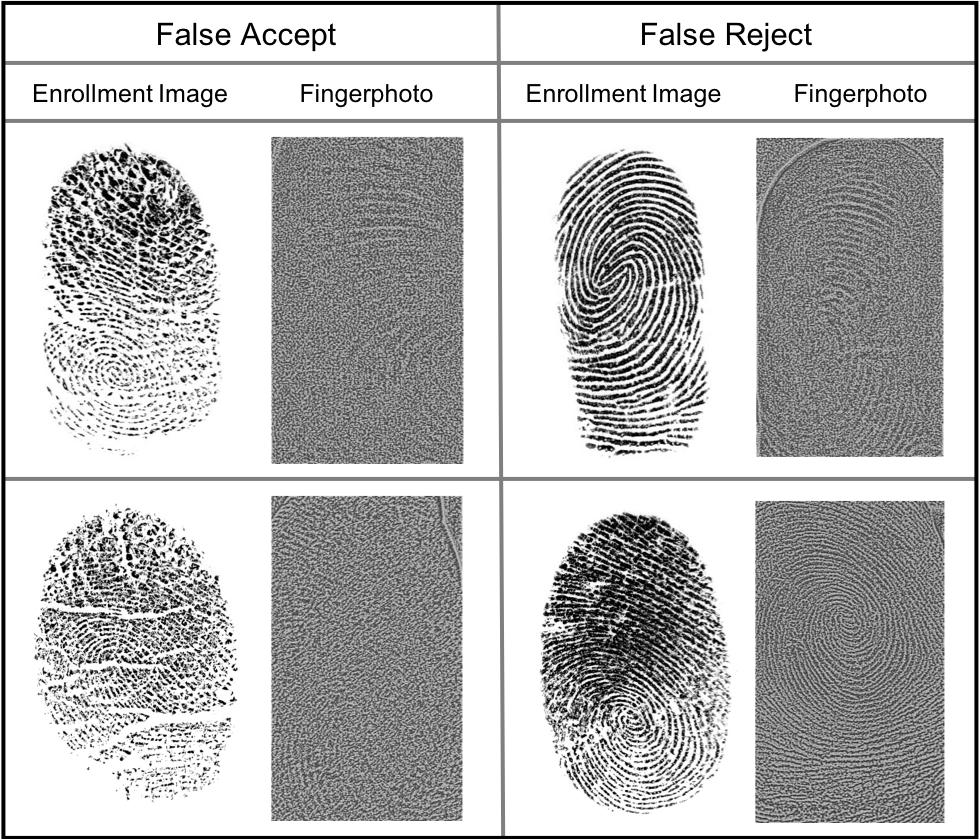}
        \caption{Example image pairs of enrollment images from contact-based slap reader and fingerphotos captured by App1, resulting in False Accepts and False Rejects.}
        \label{fig:failurecases}
\end{figure}

\begin{figure}[t!]
    \centering
        \includegraphics[width=\linewidth]{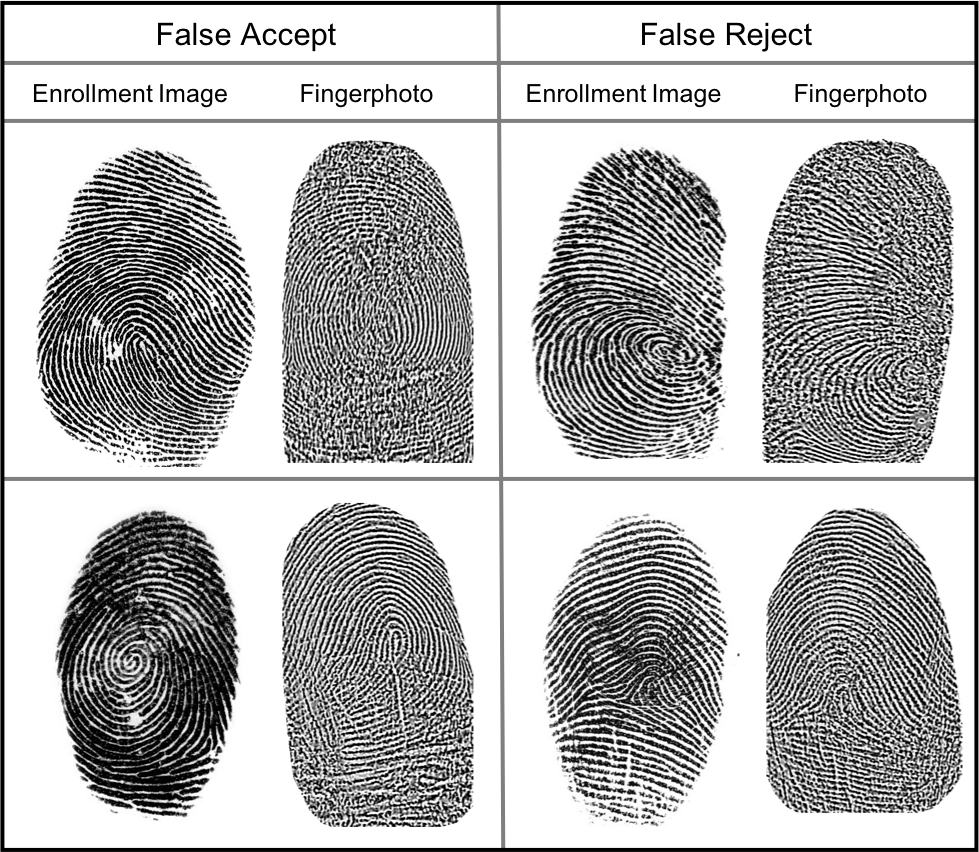}
        \caption{Example image pairs of enrollment images from contact-based slap reader and fingerphotos captured by App2, resulting in False Accepts and False Rejects.}
        \label{fig:failurecases2}
\end{figure}

 We also report the \textbf{FAR @ FRR = 2.0\%} for both apps under the four finger fusion scenario in Table \ref{tab:far_frr}. This metric of fixing the False Reject Rate (FRR) to 2\% was suggested by Caribou Digital (the sponsors for app development and evaluation) based on conversations with government officials who suggested this would be a basic hurdle rate for devices to be used in some large government systems. This metric is important since one of the major problems identified in Aadhaar program is the ``exclusion" of individuals with poor quality fingerprints from receiving their benefits. Limiting the false reject rate ensures that no deserving beneficiary is denied the due benefits. Hence low false rejects are often of higher priority than low false accepts.

\begin{table}[htbp!]
\centering
\caption{FAR (\%) @ FRR = 2.0\% reported under the four finger fusion scenario.}
\bigskip
\label{tab:far_frr}
\resizebox{0.8\linewidth}{!}{
\begin{tabular}{cccc}
\toprule
                        &   \textbf{App1} & \textbf{App2} & \textbf{SilkID} \\ \midrule
\textbf{FAR (\%) @ FRR=2\%}  &   56.20      &     0.86  &     0.00 \\ \bottomrule           
\end{tabular}
}
\end{table}


The low observed performance of App1 is most likely because App1 does not wait sufficiently long for the smartphone camera to focus the fingerprint in the field of view before capturing its image, resulting in relatively poor quality fingerprint images. See also false reject and false accept cases for the two Apps in Figs.~\ref{fig:failurecases} and ~\ref{fig:failurecases2}, and the failure to acquire rates in Table.~\ref{tab:fta}.


\subsection{Fingerphoto-to-Fingerphoto Matching}
In addition to fingerphoto-to-legacy-fingerprint matching, we also report the performance on matching fingerphoto-to-fingerphoto captured by the same App. Although, the two fingerphotos of the same finger are captured consecutively within a maximum time gap of 30 seconds, contact-less capture environment and hand movement induce large variations in the two fingerphotos. Table.~\ref{tab:fp2fp} presents the TAR $@$ FAR = 0.1\% for matching fingerphoto-to-fingerphoto of individual fingers and four finger fusion for the two Apps.

\begin{table}[!t]
    \centering
    \caption{TAR $@$ FAR = 0.1\% for the fingerphoto-to-fingerphoto matching for the apps: \textit{App1} and \textit{App2}. Here, \textit{RT}, \textit{LT}, \textit{RI}, \textit{LI}, and \textit{FF} refers to Right Thumb, Left Thumb, Right Index, Left Index, and Four Finger Fusion, respectively.}
    \vspace{1mm}
    \resizebox{0.8\linewidth}{!}{
    \begin{tabular}{lccccc}
    \toprule
        & \textbf{RT} & \textbf{LT} & \textbf{RI} & \textbf{LI} &  \textbf{FF} \\ 
        \midrule
        \textit{App1} & 46.95 & 50.76 & 27.48 & 22.15 & \textbf{72.14} \\ \hline
        \textit{App2} & 80.82 & 82.88 & 83.56 & 85.62 & \textbf{99.66} \\
         \bottomrule
    \end{tabular}
    }
    \label{tab:fp2fp}
\end{table}

\subsection{Baseline Performance}
In order to establish a baseline verification performance, we compute comparison scores between CrossMatch enrollment images and SilkID verification images. We then plot the ROC curve for the baseline as shown in Figure~\ref{fig:ind_fingers}.

\subsection{Failure Cases}

In order to gain a deeper insight into the verification performances for the two apps, we observe example images where (i) the genuine comparison scores between the enrollment images and the images from the app are very low (false rejects), and (ii) the impostor scores between the legacy enrollment images and the images from the app are relatively high (false accepts). See Fig.~\ref{fig:failurecases}. In addition, Table~\ref{tab:fta} shows the failure to acquire rates (\%) for the two apps and the two optical sensors.

\begin{table}[htbp!]
\centering
\caption{Failure to Acquire Rates (\%) for App1, App2, Crossmatch slap scanner, and SilkID fingerprint reader. The total number of acquisitions is 2,472 each for CrossMatch slap scanner, App1, and App2, and 560 for SilkID reader.}
\bigskip
\label{tab:fta}
\resizebox{0.8\linewidth}{!}{
\begin{tabular}{ccccc}
\toprule
&   \textbf{App1} & \textbf{App2} & \textbf{CrossMatch} & \textbf{SilkID} \\ 
\midrule
\textbf{FTA (\%)}  &     28.5     &  13.0   & 0.1  & 4.0  \\ 
\bottomrule
\end{tabular}
}
\end{table}

\subsection{Fingerprint Image Quality}

\begin{figure}[htbp!]
  \centering
  \includegraphics[width=\linewidth]{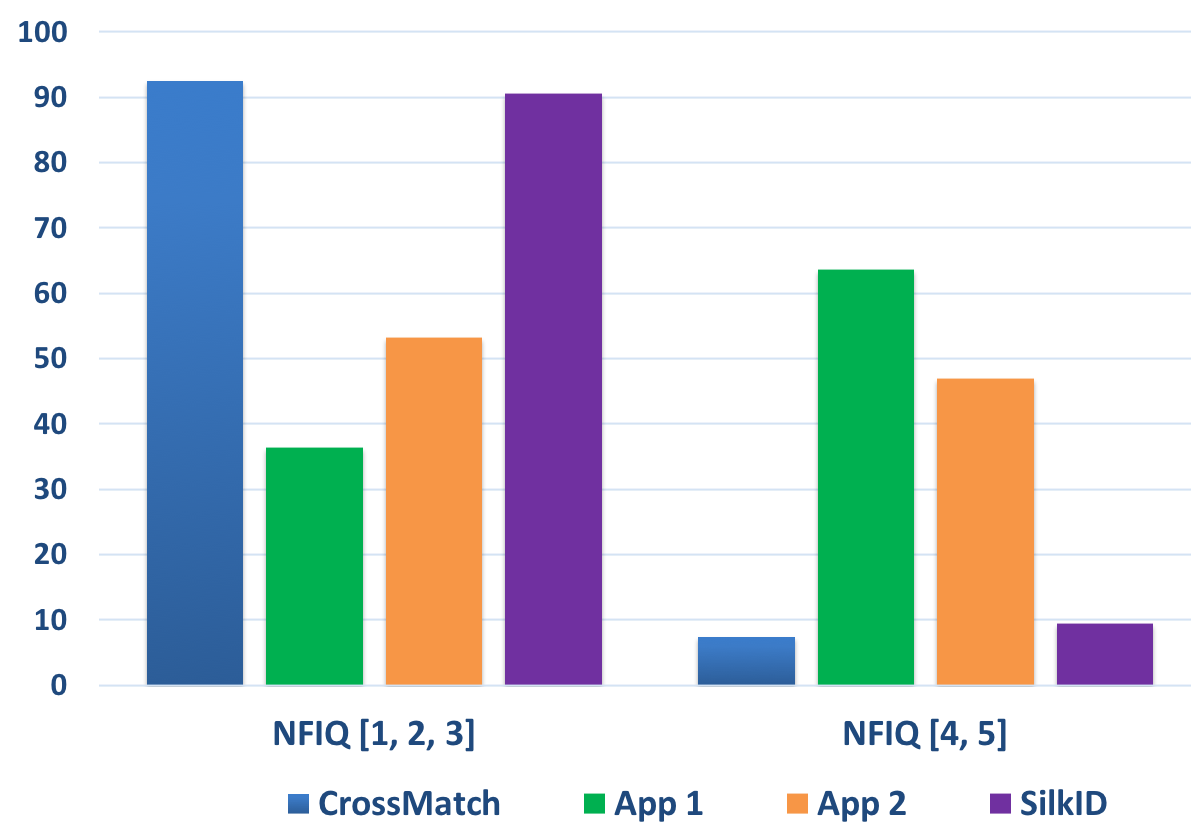}
  \caption{Histograms of NFIQ values (quantized to two levels) for fingerprint images acquired by App1, App2, Crossmatch slap scanner, and SilkID fingerprint reader.}
  \label{fig:nfiq}
\end{figure}

We also compute the fingerprint image quality based on NIST Fingerprint Image Quality (NFIQ)~\cite{nfiq}. We present histograms for the apps based on two different covariates: (i) NFIQ values in the range [1,2,3] which are images considered to be of ``good" quality, and (ii) NFIQ values in the range [4,5] which comprise of ``poor" quality fingerprint images (see Figure~\ref{fig:nfiq}).

\begin{figure}
    \centering
    \includegraphics[width=0.45\linewidth]{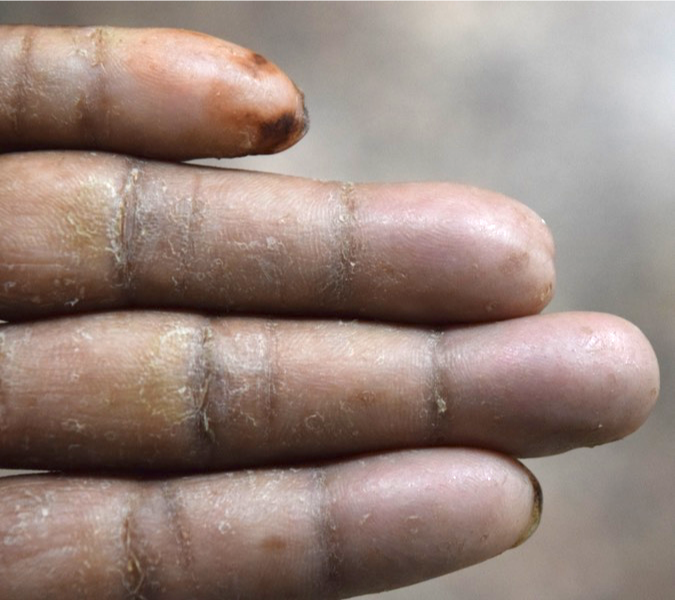}
    \includegraphics[width=0.45\linewidth]{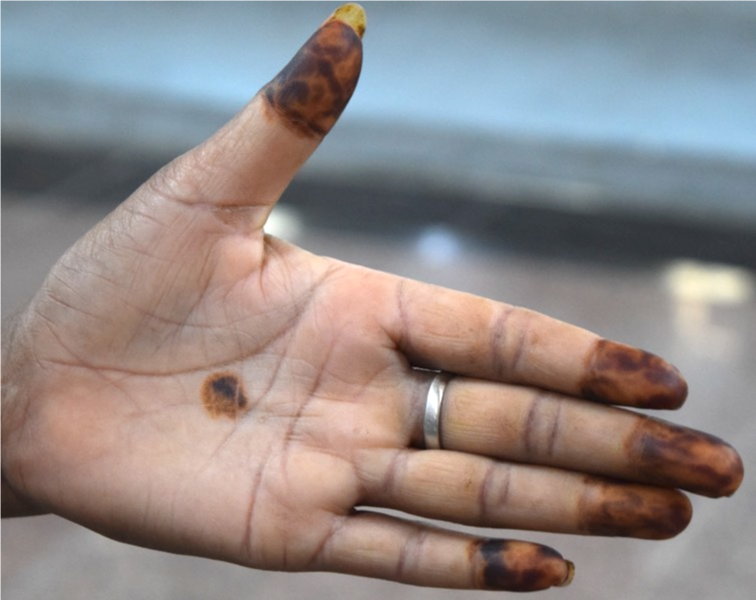}
    \caption{Some of the challenges in fingerprint acquisition of subject with (a) worn-out and damaged fingers and (b) fingers with henna.}
    \label{fig:challenges}
\end{figure}

\section{Conclusions}

This study presents data collection protocol and analysis of an \textit{in-situ} evaluation of matching fingerphotos to slap fingerprint images. A total of 309 subjects, employed in different occupations, such as construction, gardening, etc. were enrolled using a contact-based slap reader, and verified using fingerphotos captured by two contact-less Android apps installed on a commodity smartphone. A baseline performance in matching accuracy was established based on fingerprint images from a different contact-based optical reader. Experimental results show that score fusion of four fingers using sum rule is able to achieve a TAR of 95.79\% at FAR of 0.1\% for one of the apps, compared to the baseline performance of TAR = 98.55\% at FAR = 0.1\%. However, the best performance of individual fingers in matching fingerphotos to slap fingerprints is only 76.63\% TAR @ FAR = 0.1\%. This is significantly less than the acceptable requirements. Nevertheless, the accuracy of fingerphoto matching to slap images for a fusion of four fingers gives us encouragement for further developing the fingerphoto technology and both companies are currently engaged in R\&D to significantly improve performance and refine liveness detection. Future work would include (i) robustness to fingerprint acquisition challenges (as shown in Fig.~\ref{fig:challenges}), (ii) high throughput, and (iii) improved quality of captured fingerphotos.

%
%
%

{\small
\bibliographystyle{unsrt}
\bibliography{egbib}
}

\end{document}